\documentclass[conference,a4paper]{IEEEtran}

\usepackage{amsmath}
\usepackage{amssymb}
\usepackage{amsthm}
\usepackage{booktabs}
\usepackage{array}
\usepackage{enumitem}
\usepackage{hyperref}
\usepackage{xcolor}
\usepackage{listings}

\hypersetup{
  hidelinks,
  pdftitle={LegalFarePlan: A Label-Setting Framework for Fare-Transparent Urban Rail Route Planning under Non-Additive Fare Rules},
  pdfauthor={Tanghui Li},
  pdfkeywords={fare-aware route planning, urban rail transit, non-additive fares, legal exit-and-reentry, label-setting search, reproducible artifact}
}

\newtheorem{proposition}{Proposition}

\newcommand{\fare}{\mathsf{fare}}
\newcommand{\penaltyTime}{\mathsf{penalty}^{time}}
\newcommand{\penaltyMoney}{\mathsf{penalty}^{money}}
\newcommand{\shortestTime}{\mathsf{shortestTime}}
\newcommand{\dom}{\preceq}

\setlist[itemize,enumerate]{leftmargin=*,nosep}
\title{LegalFarePlan: A Label-Setting Framework for Fare-Transparent Urban Rail Route Planning under Non-Additive Fare Rules}
\author{
\IEEEauthorblockN{Tanghui Li}
\IEEEauthorblockA{Tongji University\\
Shanghai, China\\
thli@tongji.edu.cn}
}

\begin{document}
\maketitle
\thispagestyle{empty}

\begin{abstract}
Urban rail fare systems may be non-additive: the fare of a single paid journey from an origin to a destination can differ from the sum of fares over multiple legally separated journey legs. This paper presents \emph{LegalFarePlan}, a fare-transparent route-planning framework that models legal exit-and-reentry operations as explicit, auditable constraints. Given a transit network, fare function, transfer rules, station-level exit/re-entry costs, an extra-time budget, and a split limit, the planner computes explainable route plans over paid journey segments. The artifact implements Dijkstra shortest-time and direct route-planner baselines, a greedy split heuristic, bounded exact label-setting, and Pareto-frontier search. Evaluation uses controlled synthetic data and a 57-station semi-synthetic benchmark with 360 OD pairs. On the semi-synthetic benchmark, bounded exact search identifies positive modeled fare reductions for 71.11\% of OD pairs, with mean reduction 3.78 and maximum reduction 9.0 synthetic fare units under a 45-minute extra-time budget. These results demonstrate method behavior and reproducibility; they are not empirical conclusions about MTR or any transit operator.
\end{abstract}

\begin{IEEEkeywords}
fare-aware route planning; urban rail transit; non-additive fares; legal exit-and-reentry; label-setting search; reproducible artifact.
\end{IEEEkeywords}

\noindent\textit{Data and legality scope.}
All exit-and-reentry strategies in this paper are modeled as legal behavior: a passenger exits through normal gates, re-enters through normal gates, and pays the published fare for every paid leg. The artifact does not model fare evasion, ticket misuse, system tampering, gate manipulation, or regulatory avoidance. The included datasets are synthetic or semi-synthetic and are used only for algorithm validation and reproducibility.

\section{Introduction}

Transit route planning is usually framed as a shortest-path or multicriteria routing problem in which the planner optimizes travel time, transfers, walking, reliability, or generalized cost. Fare is often treated as a static OD attribute. This simplification is insufficient for transit systems with non-additive fare rules. In such systems, the fare for one continuous paid journey from station $o$ to station $d$ may differ from the total fare obtained by splitting the journey into several legal paid legs. A passenger may legally exit at an intermediate station and re-enter, but this action has time, inconvenience, and sometimes monetary costs.

This work studies the following route-planning problem:
\begin{quote}
Given a rail network, a non-additive fare table, explicit transfer rules, legal exit/re-entry costs, and user constraints, compute a route plan that minimizes total paid fare while reporting time, transfers, number of exit/re-entry operations, and a human-readable explanation.
\end{quote}

The problem is not reducible to a physical shortest path. The planner must combine two layers: a physical routing layer that checks whether each leg can be traveled through the network, and a fare layer that prices each legal paid journey. This separation is important for reproducibility: physical network data, fare tables, transfer rules, and legal exit assumptions can each be audited independently.

The contributions are:
\begin{enumerate}
  \item A formal definition of fare-aware route planning with legal exit-and-reentry operations under non-additive transit fare rules.
  \item A reproducible CSV schema covering stations, physical edges, fares, transfer rules, exit/re-entry penalties, and OD benchmarks.
  \item A standard-library Python artifact with Dijkstra shortest-time, direct route-planner, greedy, bounded exact label-setting, and Pareto-frontier modes.
  \item A controlled 8-station synthetic benchmark and a larger 57-station semi-synthetic benchmark for reproducible evaluation when licensed real data are unavailable.
  \item Experiment scripts, tests, tables, and figures that make all reported modeled fare-reduction results traceable to program output.
\end{enumerate}

\section{Related Work}

The physical routing component builds on shortest-path search, beginning with Dijkstra's algorithm \cite{dijkstra1959}. Modern public-transit journey planners extend shortest paths to timetable, transfer, and multimodal constraints \cite{bast2015,dibbelt2017,witt2015}. The proposed fare-aware problem is also related to multicriteria shortest paths, where labels are maintained over several objectives and dominated labels are pruned \cite{kurbanov2022}. However, fare-aware exit-and-reentry planning differs from standard multicriteria routing because the objective is computed over \emph{paid journey segments}, not simply over physical edges.

Transit network analysis is also connected to assignment models. Wardrop's equilibrium principles provide an early foundation for route-choice modeling \cite{wardrop1952}, while Spiess and Florian formulate optimal strategies for transit assignment under service uncertainty \cite{spiess1989}. This paper does not solve a system-level assignment problem, but it borrows the view that passenger-facing plans should be evaluated by explicit generalized costs and operational constraints.

Fare-aware routing further depends on ticketing and fare data. Smart-card and automated fare-collection data have been widely studied for OD inference and public-transit analysis \cite{barry2002,bagchi2005,pelletier2011,seaborn2009}. These studies motivate the need for careful data provenance and fare-product interpretation. Unlike demand-estimation work, LegalFarePlan treats fare tables as an explicit algorithmic input and focuses on legally explainable route plans.

This artifact also relates to reproducible computational research. Rather than reporting operator-specific conclusions without auditable evidence, the system separates data schema, validation, algorithms, and generated outputs. This is essential because transit fare tables are policy-dependent, ticket-product-dependent, and time-varying. Artifact review and badging practices similarly emphasize the importance of reusable computational evidence \cite{acm2020}.

\section{Motivating Example}

The controlled synthetic benchmark contains a compact urban rail network with three lines and eight stations. The direct synthetic fare from Alder Central (A) to Harbor Expo (H) is 18.0. The legal paid leg from A to Elm Park (E) costs 8.0, and the legal paid leg from E to H costs 5.0. If a passenger exits and re-enters at E, the total paid fare is 13.0. The physical journey remains valid, and the model adds a station-specific exit/re-entry time penalty.

The optimizer reports a legal split at Elm Park (E), modeled fare reduction 5.0, and extra modeled time 2.0 minutes relative to the direct plan. This example illustrates the central modeling issue: the path can be physically similar, but its paid-journey decomposition changes the fare. A planner must therefore reason over both network paths and fare-leg sequences.

\section{Problem Formulation}

Let $G=(V,E)$ be an urban rail network. Each station $v \in V$ has metadata, including a flag indicating whether the model permits legal exit-and-reentry at $v$. Each physical edge $e=(u,v,\ell)$ has a line identifier $\ell$ and travel time $t(e)$. Transfer rules define whether a line change from $\ell_i$ to $\ell_j$ is allowed at station $v$ and specify the corresponding transfer time.

Let $\fare(o,d)$ be a fare function for a paid journey from station $o$ to station $d$. The function may be non-additive:
\[
\fare(o,d) \ne \fare(o,x)+\fare(x,d).
\]
Let $X \subseteq V$ be the set of stations at which legal exit-and-reentry is modeled as allowed. Each $x \in X$ has a time penalty $\penaltyTime(x)$ and optional monetary penalty $\penaltyMoney(x)$.

A fare-aware route plan for OD pair $(o,d)$ is a sequence
\[
P = (s_0=o, s_1, \ldots, s_m=d),
\]
where each intermediate station $s_i$ for $1 \le i < m$ is a legal exit-and-reentry station. Each consecutive pair $(s_i,s_{i+1})$ is one paid journey leg. The total monetary cost is
\[
C(P) =
\sum_{i=0}^{m-1} \fare(s_i,s_{i+1})
+ \sum_{i=1}^{m-1} \penaltyMoney(s_i),
\]
and the total modeled time is
\[
\begin{aligned}
T(P) ={}&
\sum_{i=0}^{m-1} \shortestTime_G(s_i,s_{i+1}) \\
&+ \sum_{i=1}^{m-1} \penaltyTime(s_i).
\end{aligned}
\]
The function $\shortestTime_G$ is computed on the physical network with transfer rules.

The optimization problem is:
\[
\begin{aligned}
\min_P \quad & C(P) \\
\text{s.t.} \quad
& m-1 \le K, \\
& T(P) \le T(P_{\mathrm{direct}}) + \Delta, \\
& s_i \in X \quad \forall i \in \{1,\ldots,m-1\}, \\
& \fare(s_i,s_{i+1}) \text{ is defined} \quad \forall i, \\
& \shortestTime_G(s_i,s_{i+1}) < \infty \quad \forall i.
\end{aligned}
\]
The primary objective is monetary cost. Ties are broken by lower travel time, fewer exits, fewer transfers, and simpler explanations.

\subsection{Assumptions and Non-Goals}

The formulation intentionally separates optimization from policy interpretation. First, the fare table is assumed to define the monetary cost of a paid leg for a fixed ticket type. Second, an exit-and-reentry operation is allowed only when the station-level data explicitly permit it, and every split creates a new paid leg. Third, travel times are deterministic edge and transfer times; timetable effects, headways, crowding, and disruptions are outside the current model. Finally, the planner is advisory: it reports modeled legal assumptions and does not replace operator rules, passenger conditions, or local regulations.

\section{Algorithms}

\subsection{Physical Routing}

The physical layer runs Dijkstra-style search over station-line states $(v,\ell)$. Moving along an edge adds in-vehicle time. Switching from line $\ell_i$ to $\ell_j$ at station $v$ is allowed only when an explicit transfer rule exists and is marked valid. This design avoids name-based transfer inference, which is unsafe for reproducible transit routing.

\subsection{Baselines}

\textbf{Direct-fare baseline.} The direct baseline returns the one paid journey $(o,d)$ if $\fare(o,d)$ exists and a physical route is feasible.

\textbf{Shortest-time baseline.} The shortest-time baseline computes the physically shortest route without inserting paid-journey splits. In the current artifact, the direct and shortest-time baselines share the same paid-leg structure but differ conceptually: one defines the fare comparison point, while the other defines the physical time reference.

\subsection{Paid-Leg Search Space}

The fare layer can be viewed as a directed auxiliary graph whose nodes are stations and whose arcs are legal paid legs. An arc $(u,v)$ exists only when a fare entry $\fare(u,v)$ is available and the physical routing layer can find a feasible path from $u$ to $v$. A route plan is therefore a path in this auxiliary graph, with intermediate nodes constrained to be legal exit-and-reentry stations. This view makes the non-additive fare rule explicit: costs are attached to paid-leg arcs, not to physical track edges.

\subsection{Greedy Split Heuristic}

The greedy heuristic starts from the direct plan and repeatedly inserts one legal split station into the current sequence. For each insertion position and candidate station, the planner checks fare-leg availability, physical reachability, split count, and extra-time feasibility. It accepts the best improving insertion and stops when no insertion improves the plan or when $K$ splits have been used.

\subsection{Exact Bounded Label-Setting Search}

The exact search treats legal paid legs as transitions. A state is $(v,k)$, where $v$ is the current station and $k$ is the number of exit-and-reentry operations used. A label stores fare, time, split count, and the station sequence. A label $a$ dominates label $b$, denoted $a \dom b$, if $a$ is no worse in fare, time, and split count, and strictly better in at least one of them.

The search initializes the direct plan as an incumbent, sets the time limit to $T(P_{\mathrm{direct}})+\Delta$, and expands labels in increasing fare order. From a label at station $v$, it tests every station $x$ as either the destination or a legal split station. A transition is accepted only if the paid leg $(v,x)$ has a fare entry, the physical path is feasible, the split count is at most $K$, and the accumulated time is within the limit. Destination labels update the incumbent; intermediate labels are inserted only when they are not dominated at the same station-split state.

\begin{proposition}[Bounded optimality]
For a fixed OD pair, split limit $K$, extra-time budget $\Delta$, fare table, physical graph, transfer rules, and exit/re-entry penalty table, the exact search returns a feasible plan with minimum monetary cost among all legal paid-leg sequences that satisfy the constraints, assuming no feasible nondominated label is discarded.
\end{proposition}

\noindent\textbf{Sketch.}
The search enumerates feasible legal paid-leg transitions up to $K$ splits and the time limit. A discarded label is dominated by another label at the same state with no greater fare, time, or split count, so extending the dominated label cannot yield a better feasible plan under nonnegative leg costs and penalties. Therefore, pruning dominated labels preserves at least one representative of every potentially optimal continuation.

\begin{proposition}[Direct upper bound]
If the direct plan is feasible and included in the candidate set, the exact optimizer's returned monetary cost is no greater than the direct fare.
\end{proposition}

\noindent\textbf{Sketch.}
The algorithm initializes the incumbent with the direct plan. It replaces the incumbent only with lexicographically better feasible plans. Hence the final plan cannot have higher monetary cost than the direct incumbent.

\subsection{Pareto Frontier Search}

The Pareto mode returns nondominated feasible plans over fare, travel time, exit penalty, number of splits, and transfers. This mode is useful when a decision maker wants to inspect fare-time trade-offs rather than accept a single scalarized result.

\subsection{Complexity}

Let $n=|V|$, $K$ be the split limit, $L$ be the number of nondominated labels retained per station-split state, and $R$ be the cost of one physical shortest-path query. Without caching, checking all candidate paid legs during search can be expensive. LegalFarePlan therefore caches physical paths and paid-leg feasibility; each distinct OD leg is solved at most once on the station-line graph. The bounded fare-layer search has at most $O(nK L)$ retained labels and considers up to $O(n)$ outgoing paid-leg candidates per label, giving $O(n^2 K L)$ fare-layer transition checks after path caching. The path-cache construction is bounded by $O(n^2 R)$. In practice, $K$ is a small user-facing constraint, and fare-leg availability plus extra-time constraints reduce the search space.

\section{System Design}

The artifact is a standalone Python research package. Data loading, physical routing, optimization, reporting, experiments, table/figure generation, and tests are separated into explicit modules. For each OD pair, the output includes original route and fare, optimized route, exit/re-entry sequence, optimized fare, fare reduction, extra travel time, number of extra exits, transfers, a human-readable explanation, and legality assumptions.

The validation layer enforces several auditability invariants before experiments run: station identifiers and names must be unique; edge endpoints, fares, transfer rules, penalties, and OD pairs must reference known stations; duplicate fare entries for the same OD and ticket type must be consistent; and transfer rules may only reference lines listed at the corresponding station. These checks are deliberately simple but important, because fare-aware results are only meaningful when the input fare and network layers are internally consistent.

\section{Dataset and Reproducibility}

\subsection{Synthetic Dataset}

The minimal controlled dataset is explicitly synthetic. It contains 8 stations, 3 lines, 7 bidirectional physical edges, station-level exit/re-entry penalties, a complete synthetic adult fare table, and 10 OD benchmark pairs. It is intended to test algorithmic behavior and reproducibility, not to characterize any real transit system.

\subsection{Semi-Synthetic Urban Rail Benchmark}

To evaluate scalability beyond the minimal controlled dataset, the artifact also includes \texttt{data/semi\_synthetic\_rail/}. This dataset is deterministic and semi-synthetic: it is not copied from MTR or any official operator source. It contains 57 stations, 5 lines, 4 interchange hubs, 58 bidirectional physical edges, 3249 adult fare entries, station-level exit/re-entry penalties, explicit transfer rules, and 360 OD benchmark pairs.

The construction follows common structural features of urban rail systems: line corridors are connected through transfer hubs, stations are assigned coarse fare zones, and fares depend on shortest physical travel time, zone difference, and non-additive cross-corridor surcharges. The fare generator computes a shortest physical time $\tau(o,d)$, applies a base fare, a time component, a zone-difference component, and deterministic surcharge terms for selected cross-corridor OD pairs, and then rounds fares to half-unit synthetic currency. In the released generator, the pre-rounding fare is
\[
2.8 + 0.16\tau(o,d) + 0.75|\mathrm{zone}(o)-\mathrm{zone}(d)| + \sigma(o,d),
\]
where $\sigma(o,d)$ encodes the non-additive cross-corridor and long-trip surcharges. The surcharge terms are included to stress-test fare transparency under non-additive rules. Because the dataset is generated by \texttt{scripts/generate\_semisynthetic\_data.py}, all topology, transfer, penalty, and fare assumptions are reproducible and inspectable.

The required CSV files are \texttt{stations.csv}, \texttt{edges.csv}, \texttt{fares.csv}, \texttt{transfer\_rules.csv}, \texttt{exit\_reentry\_penalty.csv}, and \texttt{od\_pairs.csv}; they respectively encode station metadata, physical edges, fare entries, transfer permissions, legal exit/re-entry penalties, and benchmark OD pairs.

\subsection{Real-Data Requirements}

Any operator-specific deployment, including an MTR case study, must add source URLs, license terms, retrieval dates, preprocessing scripts, fare-product definitions, special fare rules, and a legal review of exit-and-reentry assumptions before reporting real-world fare-reduction results.

\section{Experimental Setup}

The 8-station synthetic benchmark is used as a minimal verification benchmark and motivating example with split limit $K=2$ and maximum extra time $\Delta=30$ minutes. The larger semi-synthetic benchmark is the main experiment and is run with synthetic adult fares, split limit $K=2$ unless varied, maximum extra time $\Delta=45$ minutes, and station-specific legal exit/re-entry time penalties.

We compare five modes: Dijkstra shortest-time routing without paid-journey splits, a direct route-planner baseline using one paid OD fare, greedy split insertion, bounded Pareto frontier search, and bounded exact label-setting search. We report modeled fare-reduction distribution, top examples, percentage of OD pairs with positive reductions, trade-off between fare and extra time, sensitivity to $K$, sensitivity to exit/re-entry penalty, comparison against baselines, and runtime scaling over sampled sub-networks.

\subsection{Metrics}

For a feasible OD pair, \emph{fare reduction} is the direct one-paid-journey fare minus the optimized fare; it is reported in synthetic fare units. \emph{Extra time} is the optimized modeled time minus the direct modeled time, including exit/re-entry penalties. \emph{Positive OD} is the percentage of feasible OD pairs with strictly positive fare reduction. \emph{Gap OD} counts OD pairs for which a baseline's fare is higher than the bounded exact result. Runtime scaling measures wall-clock time for exact search over deterministic OD samples at each network size and split limit.

\section{Results}

\textbf{Scope.} All results in this section are generated from synthetic or semi-synthetic data. They validate method behavior and reproducibility; they do not describe MTR fares, passenger outcomes, or any official transit policy.

\begin{table*}[t]
\centering
\caption{Benchmark summary. Fare units are synthetic; reductions are computed against the direct one-paid-journey strategy.}
\label{tab:summary}
\begin{tabular}{lrrrrrr}
\toprule
Dataset & Stations & OD pairs & Feasible & Positive OD & Mean reduction & Max reduction \\
\midrule
Controlled synthetic & 8 & 10 & 10 & 80.00\% & 2.70 & 5.00 \\
Semi-synthetic rail & 57 & 360 & 360 & 71.11\% & 3.78 & 9.00 \\
\bottomrule
\end{tabular}
\end{table*}

Table~\ref{tab:summary} shows that the engine scales from the 8-station controlled benchmark to the 57-station semi-synthetic benchmark. On the larger benchmark, 256 of 360 OD pairs have positive modeled fare reductions. The mean extra travel time is 1.19 minutes, with observed extra time ranging from -1.5 to 4.5 minutes under the modeled shortest-path and penalty assumptions.

\begin{table}[t]
\centering
\caption{Average baseline comparison on the 57-station semi-synthetic benchmark.}
\label{tab:baseline}
\begin{tabular}{lrrr}
\toprule
Method & Mean fare & Mean time & Gap OD \\
\midrule
Dijkstra/direct & 16.10 & 43.13 & -- \\
Greedy split & 12.31 & 44.32 & 0 \\
Pareto frontier & 12.31 & 44.32 & 0 \\
Bounded exact & 12.31 & 44.32 & -- \\
\bottomrule
\end{tabular}
\end{table}

Table~\ref{tab:baseline} compares route-planning modes. In this generated benchmark, greedy and Pareto-best plans match bounded exact search for all 360 OD pairs, so the gap count is zero. This should not be interpreted as a general guarantee for greedy insertion; it reflects the constructed fare rules and $K=2$ setting. The Pareto frontier contains 3.43 nondominated plans on average and at most 10 plans.

The top modeled reductions are B09--C04, B09--C06, and B09--C09, each with reduction 9.0 and one legal split at H4. The next two are B01--D09 and B02--C04, with reduction 8.5 and legal splits at H4 and H1, respectively. Each listed plan pays a separate fare for each legal paid leg.

\begin{table}[t]
\centering
\caption{Sensitivity to maximum legal exit-and-reentry operations on the semi-synthetic benchmark.}
\label{tab:split-sensitivity}
\begin{tabular}{rrrrr}
\toprule
$K$ & Positive OD & Mean red. & Max red. & Runtime \\
\midrule
0 & 0.00\% & 0.00 & 0.00 & 0.023 s \\
1 & 71.11\% & 3.78 & 9.00 & 0.277 s \\
2 & 71.11\% & 3.78 & 9.00 & 1.847 s \\
\bottomrule
\end{tabular}
\end{table}

Table~\ref{tab:split-sensitivity} confirms that $K=0$ recovers the direct strategy. In this benchmark, one legal exit is sufficient to expose all positive reductions; allowing two exits increases runtime without improving the objective.

Uniform exit/re-entry penalty changes the time trade-off but not the selected fare objective under the 45-minute budget: at 0, 3, 6, 9, and 12 minutes, positive OD remains 71.11\%, mean reduction remains 3.78, and mean extra time changes from -3.27 to -1.13, 1.00, 3.13, and 5.27 minutes. Negative extra time is possible because the fare-aware decomposition may choose a physically shorter paid-leg combination than the direct one-paid-journey baseline.

\begin{table*}[t]
\centering
\caption{Runtime scaling on semi-synthetic sub-networks. Times are seconds for sampled OD pairs.}
\label{tab:runtime}
\begin{tabular}{rrrrrr}
\toprule
Stations & OD sample & Solved & $K=0$ & $K=1$ & $K=2$ \\
\midrule
12 & 38 & 38 & 0.0006 & 0.0036 & 0.0114 \\
24 & 171 & 171 & 0.0038 & 0.0306 & 0.1653 \\
36 & 407 & 381 & 0.0113 & 0.1026 & 0.6971 \\
48 & 731 & 641 & 0.0248 & 0.2509 & 2.0067 \\
57 & 900 & 900 & 0.0384 & 0.4725 & 4.2641 \\
\bottomrule
\end{tabular}
\end{table*}

Table~\ref{tab:runtime} shows the expected growth with network size and split limit. The 36- and 48-station sub-network samples are partial prefixes of the generated network and therefore contain some disconnected OD pairs; the full 57-station benchmark is connected for all sampled OD pairs.

\subsection{Interpretation}

The semi-synthetic results show three properties of the formulation. First, the direct Dijkstra-style route planner is not fare-aware when fare is defined over paid journeys rather than physical edges. Second, the Pareto frontier remains small on this benchmark, which makes interactive inspection plausible: users or analysts can compare a few nondominated alternatives rather than a single opaque recommendation. Third, the split limit is an operationally meaningful control parameter. In the generated benchmark, one legal exit is sufficient for all positive reductions, while increasing $K$ from 1 to 2 primarily increases search cost. This illustrates why split limits and extra-time budgets should be exposed as explicit planning constraints.

\section{Case Study}

For semi-synthetic OD pair B09 to C04, the direct route passes H3, H4, and H2 with one paid journey. LegalFarePlan instead decomposes the paid journey at University Interchange (H4), producing two legal paid legs. The direct modeled fare/time are 24.5 and 63.0 minutes; the optimized modeled fare/time are 15.5 and 64.5 minutes. Thus the fare reduction is 9.0 synthetic units with 1.5 minutes of extra modeled time and one legal exit. The generated explanation states that each leg is charged separately and that the H4 exit/re-entry penalty is included in travel time.

\section{Threats to Validity and Limitations}

\textbf{External validity.} The benchmarks are synthetic or semi-synthetic; they support algorithm validation rather than empirical conclusions about a real operator. Full MTR results require licensed data, current fare rules, and audited preprocessing.

\textbf{Model validity.} The artifact does not model headways, timetables, crowding, service disruptions, comfort, station closures, or passenger acceptance. Exit/re-entry costs are station-level constants.

\textbf{Fare-policy validity.} Real fare systems may include special ticket products, time windows, transfer discounts, promotions, concessions, Airport Express rules, and operator-specific restrictions. These must be represented explicitly before deployment.

\textbf{Algorithmic validity.} The bounded exact search is exact only within the modeled constraints and the provided fare/route data. If the data omit legal constraints, the optimizer may produce legally invalid recommendations.

\section{Ethical and Legal Considerations}

The optimizer is designed for legal fare transparency. It should not be framed as a way to evade fares or bypass policy. Any user-facing system should state that every split requires a legal exit, legal re-entry, and payment of each leg's fare. A real deployment should consult current operator terms and local regulations before publishing recommendations.

\section{Conclusion}

Fare-aware route planning under non-additive fare rules is a constrained optimization problem over physical routes and paid journey segments. This paper presents LegalFarePlan, a reproducible artifact with a formal problem definition, auditable data schema, legal strategy assumptions, exact and heuristic algorithms, explainable outputs, tests, and synthetic/semi-synthetic experiments. The evaluation demonstrates that the engine can generate feasible, explainable, and legally constrained fare-aware plans on controlled benchmarks. The same pipeline can be extended to licensed full-network data when such data and policy assumptions are available.

\appendices
\section{Artifact Appendix}

The artifact folder contains the paper, data, scripts, and package code. The reproducibility script regenerates the semi-synthetic data, validates both datasets, runs tests, executes optimization, compares baselines, and regenerates tables and figures.

\section{Checklist for a Real MTR Study}

To turn this artifact into a real MTR study, the dataset must be replaced by licensed CSV inputs with documented sources, licenses, retrieval dates, preprocessing scripts, ticket-product rules, special fares, discounts, and legally validated exit/re-entry assumptions. Real results should be reported only after all OD pairs are regenerated from that audited data pipeline.

\end{document}